\documentclass[lettersize,journal]{IEEEtran}
\usepackage{amsmath,amsfonts}
\usepackage{algorithm}
\usepackage{algorithmic}
\usepackage{array}
\usepackage[caption=false,font=normalsize,labelfont=sf,textfont=sf]{subfig}
\usepackage{textcomp}
\usepackage{stfloats}
\usepackage[hyphens]{url}
\usepackage{verbatim}
\usepackage{graphicx}
\usepackage{multirow}
\usepackage{booktabs}
\usepackage{tabularx}
\usepackage{cite}
\usepackage[flushleft]{threeparttable}

\begin{document}

\title{ConPET: Continual Parameter-Efficient Tuning for Large Language Models}

\author{Chenyang Song, Xu Han, Zheni Zeng, Kuai Li, Chen Chen, Zhiyuan Liu, Maosong Sun and Tao Yang
\IEEEcompsocitemizethanks{\IEEEcompsocthanksitem Chenyang Song, Xu Han, Zheni Zeng, Zhiyuan Liu, and Maosong Sun are with the Department of Computer Science and Technology, Tsinghua University, Beijing 100084, China. \IEEEcompsocthanksitem Kuai Li, Chen Chen, and Tao Yang are with the Machine Learning Platform Department, Tencent, Beijing 100193, China.\protect\\
E-mail: scy22@mails.tsinghua.edu.cn,  liuzy@tsinghua.edu.cn
\IEEEcompsocthanksitem Zhiyuan Liu is the corresponding author.}}





\maketitle

\begin{abstract}

Continual learning necessitates the continual adaptation of models to newly emerging tasks while minimizing the catastrophic forgetting of old ones. This is extremely challenging for large language models (LLMs) with vanilla full-parameter tuning due to high computation costs, memory consumption, and forgetting issue. Inspired by the success of parameter-efficient tuning (PET), we propose Continual Parameter-Efficient Tuning (ConPET), a generalizable paradigm for continual task adaptation of LLMs with task-number-independent training complexity. ConPET includes two versions with different application scenarios. First, Static ConPET can adapt former continual learning methods originally designed for relatively smaller models to LLMs through PET and a dynamic replay strategy, which largely reduces the tuning costs and alleviates the over-fitting and forgetting issue. Furthermore, to maintain scalability, Dynamic ConPET adopts separate PET modules for different tasks and a PET module selector for dynamic optimal selection. In our extensive experiments, the adaptation of Static ConPET helps multiple former methods reduce the scale of tunable parameters by over 3,000 times and surpass the PET-only baseline by at least 5 points on five smaller benchmarks, while Dynamic ConPET gains its advantage on the largest dataset. The codes and datasets are available at \url{https://github.com/Raincleared-Song/ConPET}.

\end{abstract}

\begin{IEEEkeywords}

Continual learning, parameter-efficient tuning, large language models.

\end{IEEEkeywords}

\section{Introduction}

\IEEEPARstart{R}{ecently}, large language models (LLMs) have shown excellent capabilities from a wide range of aspects~\cite{brown2020language,wei2021finetuned,ouyang2022training}, which equip them with larger potential in handling various task-specific settings. To adapt LLMs to downstream tasks, fine-tuning is often the first choice~\cite{ding2022delta}. However, in real-life applications, the consistent emergence of materials such as the latest corpus~\cite{jin2022lifelong}, new knowledge~\cite{daruna2021continual,monaikul2021continual}, and heterogeneous tools~\cite{qin2023tool} can frequently change the task schemas. This necessitates continual task-specific adaptation of LLMs, which is highly expensive and performance-risky when conducted through traditional fine-tuning due to the huge number of LLM parameters and the catastrophic forgetting issue~\cite{zhao2022consistent}, namely the significant performance decrease on old tasks after adapted to new ones.

\begin{figure}[t]
    \centering
    \includegraphics[width=1\linewidth]{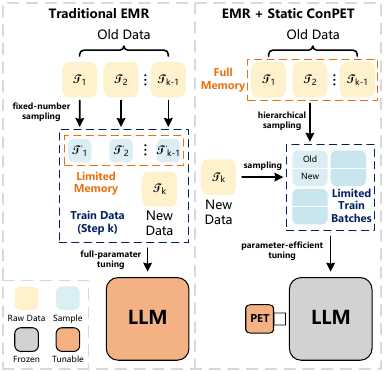}
    \caption{The comparison between traditional EMR~\cite{parisi2019continual} and adapted EMR with Static ConPET. The latter adopts the dynamic replay strategy for training data generation and PET for LLM tuning.}
    \label{fig:model-static}
\end{figure}

Although many continual learning methods have been proposed to handle these problems, specific challenges hinder their adaptation to LLMs. Dynamic-architecture-based methods~\cite{rusu2016progressive,gu2021transformer}, which progressively increase the model scale with the growth of data, suffer from unacceptable linearly growing costs due to the unrestricted scaling of the architecture. Meanwhile, most memory-based methods~\cite{rolnick2019experience,wang2019sentence,han2020continual,zhao2022consistent} frequently re-tune the model through a fixed replay strategy, where examples saved in limited episodic memory are replayed a fixed number of times with new data combined. This makes them susceptible to low scalability and over-fitting on memorized examples~\cite{han2020continual}. Furthermore, since their backbones are relatively small in scale such as BERT~\cite{devlin2018bert} and RoBERTa~\cite{liu2019roberta}, they adopt full-parameter tuning in their original works, which imposes a heavy burden on computation resources for LLMs in terms of time and GPU memory.

Faced with these challenges, we propose \textbf{Con}tinual \textbf{P}arameter-\textbf{E}fficient \textbf{T}uning (\textbf{ConPET}), a generalizable paradigm for the continual fine-tuning-based task adaptation of LLMs with task-number-independent training complexity, including \textbf{Static ConPET} and \textbf{Dynamic ConPET}.

\IEEEpubidadjcol

As shown in Figure~\ref{fig:model-static}, we first use \textbf{Static ConPET}, a general approach to adapting traditional memory-based continual methods such as EMR~\cite{parisi2019continual} to LLMs while coping with high training costs, over-fitting, and catastrophic forgetting. Specifically, it contains two key adaptations: (1) Replacing vanilla fine-tuning with parameter-efficient tuning (PET). As PET only updates tiny-scale tunable modules (generally account for lower than $1\%$ of the LLM) while keeping the original LLM frozen, the computation costs for parameter updates and the GPU memory consumption can be largely reduced~\cite{ding2022delta}. (2) Utilizing the historical data through a dynamic replay strategy, which conducts robust sampling hierarchically from full-volume historical data rather than a limited memory to improve data coverage and alleviate over-fitting and forgetting. A restriction on the number of sampled batches is also introduced to control the training complexity.

\begin{figure*}[ht]
    \centering
    \includegraphics[width=1\linewidth]{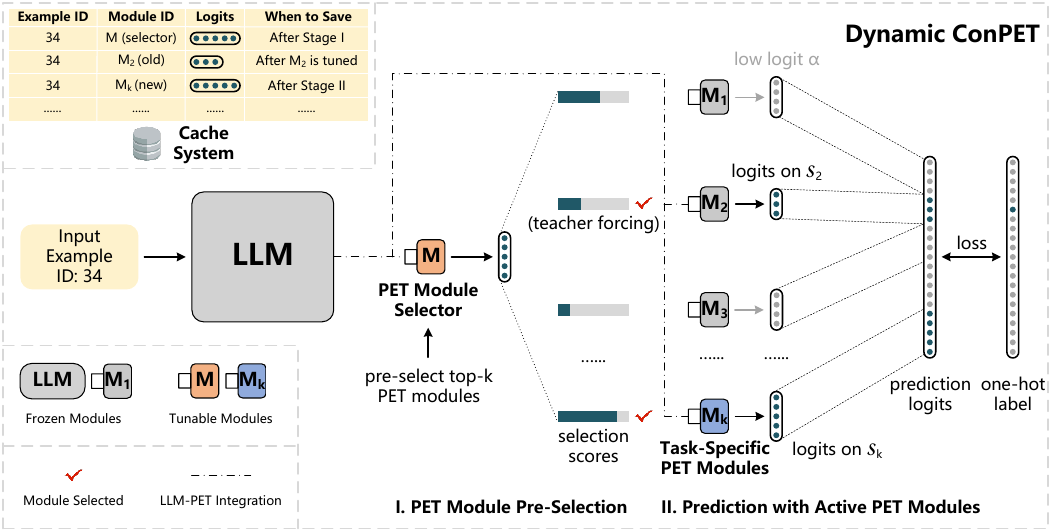}
    \caption{The architecture of ConPET when the number of active PET modules is 2. The working process can be split into two procedures: PET module pre-selection and prediction with active PET modules. All logits generated by a specific PET module will be saved instantly by the cache system after this module completes tuning.}
    \label{fig:model-large}
\end{figure*}

Adapted from memory-based methods, Static ConPET with a single PET module cannot avoid the low scalability issue. Therefore, we further propose \textbf{Dynamic ConPET} shown in Figure~\ref{fig:model-large}, a novel dynamic architecture including a backbone LLM, a PET module selector, a set of task-specific PET modules, and a cache system. Similar to mixture-of-expert (MoE) architectures, Dynamic ConPET separates the parameters for tasks with different schemas (e.g., distinct types of knowledge) in different modules, which naturally alleviates forgetting and maintains scalability to the increasing task number. As each task-specific module only contains a lightweight and pluggable PET module rather than heavy sub-networks in former dynamic-architecture-based methods, Dynamic ConPET is more tailored for memory-consuming LLMs. Meanwhile, the PET module selector ensures a constant forward propagation cost by pre-selecting a fixed quantity of task-specific modules with the highest scores to participate in the prediction.

We conduct comprehensive experiments on multiple datasets of knowledge extraction, a representative continual learning scenario with consistently emerging new knowledge types. The results demonstrate that both versions of ConPET are effective in the continual task-specific adaptation of LLMs while having different application scenarios. Further analysis shows the effect of parameter-efficient tuning, PET module pre-selection, and different task splits.

\section{Related Work}

\subsection{Parameter-Efficient Tuning}

To fine-tune LLM more efficiently, parameter-efficient tuning is proposed~\cite{ding2022delta}, which mainly consists of three types: (1) Addition-based methods~\cite{houlsby2019parameter,li2021prefix,gao2021making} introduce additional small-scale tunable parameters while freezing the original LLM. (2) Specification-based methods~\cite{lee2019would,zhao2020masking,zaken2022bitfit} selectively optimize part of the LLM parameters with the remaining parameters unchanged. (3) Reparameterization-based methods~\cite{hu2021lora,qin2021exploring} convert adaptive parameters of LLMs into parameter-efficient forms during optimization. In this work, ConPET can be adapted to addition-based methods and some reparameterization-based methods. Notably, LoRA~\cite{hu2021lora}, which introduces additional parameters to model the weight differences, is proven to perform better than most mainstream PET methods and thus widely adopted~\cite{ding2022delta}. Therefore, our experiments in this work will focus on LoRA as a representative.

PET is demonstrated to save considerable computation costs and memory consumption. According to previous works~\cite{sun2023comparative}, given the same instruction data and GPUs, the time consumed by PET tuning on LLaMA-7B~\cite{touvron2023llama} is only around one-fourth that of full-parameter tuning.

\subsection{Continual Learning for LLMs}

Continual learning aims at teaching a model to incrementally handle newly emerging tasks while mitigating the catastrophic forgetting issue. Despite the success of in-context learning in zero/few-shot leaning~\cite{brown2020language}, fine-tuning is still a prevalent paradigm in task adaptation of LLMs~\cite{ding2022delta}. The existing efforts on tuning-based continual learning can be generally classified into three categories: (1) Consolidation-based methods protect the parameters of importance from shifting considerably, which is often implemented through regularization~\cite{kirkpatrick2017overcoming,lee2017overcoming,chaudhry2018riemannian} or distillation~\cite{li2017learning,zhang2020class}. However, they perform poorly for lack of historical data utilization. (2) Dynamic-architecture-based methods~\cite{chen2015net2net,rusu2016progressive,gu2021transformer} maintain a model whose scale is progressively increasing with the task number. By introducing independent parameters for different tasks, they can effectively overcome catastrophic forgetting but suffer from the linear growth of training costs. (3) Memory-based methods~\cite{isele2018selective,rolnick2019experience,wang2019sentence,han2020continual,zhao2022consistent} introduce episodic memory to store and replay examples of old tasks. Although the old task information is partly retained in memory, they are susceptible to over-fitting and low scalability caused by the frequent re-training on a fixed model architecture through the fixed replay strategy. All the above methods adopt vanilla fine-tuning due to the small scale of their backbones.

Considering the expensive costs of tuning LLMs, some works integrate parameter-efficient tuning with continual learning but still cannot fully handle some shortcomings of former methods. Both LFPT5~\cite{qin2021lfpt5} and Progressive Prompts~\cite{razdaibiedina2022progressive} propose to continuously introduce and train new soft prompts for a new task to tackle catastrophic forgetting, while LFPT5 additionally generate pseudo examples for experience replay. However, similar to the aforementioned dynamic-architecture-based methods, the continual accumulation of new prompts can cause the scalability issue. AdapterCL~\cite{madotto2021continual} learns a separate Adapter~\cite{houlsby2019parameter} for each task and selects the task-specific Adapter based on the perplexity. Such a selection method causes a linear increase in forwarding costs with respect to the task number. LAE~\cite{gao2023unified} iteratively trains and ensembles two experts favored by novel/historical tasks respectively, which is susceptible to over-fitting given a fixed number of experts.

Moreover, there are also works focused on the continual pre-training problem~\cite{qin2022elle,jin2022lifelong,ke2023continual}, but they are too expensive to be applied to task-specific LLM fine-tuning, which is often conducted frequently with low resources.

\subsection{Mixture-of-Expert for LLMs}

Another line of work similar to Dynamic ConPET is the MoE architecture composed of multiple separate networks, which learn to handle different subsets of input examples or take distinct responsibilities. MoE is first demonstrated to be effective in deep learning by introducing an MoE layer stacked between LSTM modules~\cite{shazeer2017outrageously}. Later, GShard~\cite{lepikhin2020gshard}, BASELayer~\cite{lewis2021base}, HashLayer~\cite{roller2021hash}, Switch Transformers~\cite{fedus2022switch}, MoEfication~\cite{zhang2022moefication}, and DEM{\small IX}~\cite{gururangan2022demix} attempt to explore the implementation and training strategies of MoE in Transformer-based models. However, these works have to modify specific structures (e.g., the FFN layer) in Transformers, which cannot be easily adapted to the continual fine-tuning of an already pre-trained LLM. Some recent works have proved that LLMs with MoE are able to obtain supreme performances when combined with instruction tuning~\cite{shen2023mixtureofexperts,zadouri2023pushing}, illustrating the potential of such kinds of architecture.

To improve efficiency while alleviating over-fitting and catastrophic forgetting in the continual task-specific adaptation of LLMs, Static ConPET adapts memory-based continual learning methods using PET and dynamic replay strategy. Furthermore, Dynamic ConPET handles low scalability through a dynamic structure with task-specific PET modules and a selector. Unlike former MoE architectures, Dynamic ConPET is better suited for LLM tuning as each expert is a lightweight and highly pluggable PET module, which can be tuned without altering the original LLM structure or parameters. Finally, both ConPET versions offer task-number-independent training complexity.

\section{Proposed Methods}

\subsection{Task Definition}

Continual fine-tuning of LLMs aims at teaching the LLM to simultaneously handle a sequence of tasks in a specific application scenario with consistently emerging materials. We denote the set of schemas of the materials handled in the $k$-th task as $\mathcal{S}_k$, with a corresponding training set $\mathcal{T}_k$ and an evaluation set $\mathcal{Q}_k$. The schemas of different tasks are disjoint. At the $k$-th step, given the seen training data $\tilde{\mathcal{T}}_k=\cup_{i=1}^k\mathcal{T}_i$, the model is required to obtain satisfactory results on the evaluation set of the new task as well as all the $k-1$ historical tasks, namely $\tilde{\mathcal{Q}}_k=\cup_{i=1}^k\mathcal{Q}_i$ with the schema set $\tilde{\mathcal{S}}_k=\cup_{i=1}^k\mathcal{S}_i$.

Taking knowledge extraction as a representative, $\mathcal{S}_k$ is a specific subset of knowledge types (e.g., entity types or relation types). Each example in the dataset consists of an input sentence and a groud-truth label, indicating the type of knowledge expressed in the input. The LLM is then required to predict the knowledge type label with steadily satisfying accuracy along the task sequence.

\subsection{Static ConPET}

Static ConPET is a generalizable approach to adapting former memory-based continual learning methods to the continual task-specific adaptation of LLMs, which mainly consists of two parts: the PET adaptation and the dynamic replay strategy.

\subsubsection{PET Adaptation and Example Encoding}

As most former memory-based methods mainly concern relatively small-scale models~\cite{parisi2019continual,wang2019sentence,zhao2022consistent}, we first replace the vanilla fine-tuning with PET when applying them to LLMs. Specifically, instead of tuning all the parameters in LLMs, we only optimize one tiny PET module, while the LLM and remaining modules stay frozen. Considering the small size of tunable parameters, the computation costs for parameter updates and the GPU memory consumption will be significantly reduced.

\begin{table}[ht]
  \centering
  \setlength{\tabcolsep}{3.5mm}{
  \begin{threeparttable}
  \caption{The settings of the two continual knowledge extraction tasks involved in our experiments.}
  \begin{tabular}{l}
    \toprule
    \textbf{Entity Typing} \\
    \midrule
    \textbf{Target}: Determine the type of \textit{entity} contained in the input example. \\
    \textbf{Entity Markers}: \texttt{[E1]} \textit{entity} \texttt{[/E1]} \\
    \textbf{Prompt Template}: In this sentence, \textit{entity} is a \texttt{[MASK]}. \\
    \midrule
    \textbf{Relation Extraction} \\
    \midrule
    \textbf{Target}: Determine the relation between \textit{head entity} and \textit{tail entity} \\expressed in the input example. \\
    \textbf{Entity Markers}: \texttt{[E1]} \textit{head entity} \texttt{[/E1]}, \texttt{[E2]} \textit{tail entity} \texttt{[/E2]} \\
    \textbf{Prompt Template}: In this sentence, \textit{tail entity} is the \texttt{[MASK]} of \\\textit{head entity}. \\
    \bottomrule
  \end{tabular}
  \begin{tablenotes}
      \item Entity markers and prompt templates appended at the end of an input example are also displayed.
  \end{tablenotes}
  \label{tab:marker-prompt}
  \end{threeparttable}
 }
\end{table}

With the aid of PET adaptation, we can conduct more efficient example encoding, which is aimed at generating informative representations for downstream tasks. We take knowledge extraction, including entity typing and relation extraction, as a representative. To improve the quality of representations, we adopt an LLM as our backbone encoder and enhance the inputs through entity markers and prompt templates. The LLM has obtained large amounts of knowledge through unsupervised training and can convert inputs into informative hidden state vectors. Following the previous work~\cite{soares2019matching}, we apply entity markers surrounding each entity in the input example to insert entity positional information. Besides, inspired by the success of prompt tuning~\cite{ding2021prompt,lester2021power}, we append prompt templates at the end of inputs and take the hidden state of \texttt{[MASK]} as the example representation. The entity markers and prompt templates used in the two knowledge extraction tasks involved are shown in Table~\ref{tab:marker-prompt}.

Formally, given an input example $\textbf{x}$, the LLM integrated with a PET module $\mathrm{M}$ first encodes $\textbf{x}$ into its example representation, which is then projected to the corresponding logits by the linear head contained in $\mathrm{M}$. We hereinafter denote this process as follows,

\begin{equation}
    \label{eq:encoding}
    \mathbf{s} = f(\mathrm{M}, \textbf{x})
\end{equation}

\noindent
where $f$ and $\mathbf{s}$ stand for the encoding function and the logit vector respectively.

\subsubsection{Dynamic Sampling Strategy} \label{sec:dynamic-replay}

Rather than conducting replays on limited memorized examples as existing memory-based methods, ConPET utilizes historical data through a dynamic replay strategy to avoid over-fitting and control the overall training steps. Specifically, we remove the limits to storage space to improve data coverage. Instead, the replay examples are dynamically selected from the full-volume data, under a restriction on the maximum batch number at each step, which ensures task-number-independent complexity.

This strategy may challenge a common assumption in continual learning that the memory is limited and thus it is unrealistic to save full-volume data~\cite{wang2019sentence}. However, we consider it more reasonable to focus on the restriction of training costs (complexity) rather than memory in terms of price and consumption. Generally, current training corpora do not exceed the TB-level even for GPT-3 175B~\cite{brown2020language}, one of the largest existing language models. The cost of storing such scale data is much less than a single V100 GPU, let alone that training modern LLMs often requires hundreds or even thousands of GPUs~\cite{brown2020language,touvron2023llama}. More supporting facts for this claim are provided in Appendix~\ref{sec:time_memory}.

Another problem of full-data storage is statistically inefficient sampling~\cite{wang2019sentence}. To overcome this issue, we adopt hierarchical sampling when utilizing stored historical data to ensure equal coverage for the examples of each old task. Specifically, rather than direct random sampling, we first generate an old task ID and then select an example from the sub-dataset of that task. Besides, a fixed ratio is kept between old and new examples in each training batch. In this way, ConPET is more robust to the data imbalance issue and statistically more efficient in terms of equal coverage for each historical task.

\subsection{Dynamic ConPET}

Despite the efficiency of Static ConPET, there still exists a potential problem of low scalability. Specifically, under downstream tasks with extremely abundant emerging materials, the volume of knowledge to be acquired may exceed the capacity of tunable parameters and thus the performance will decrease. This issue can be further exacerbated by PET due to the tiny scale of PET modules.

Therefore, we introduce Dynamic ConPET to address this issue, which is composed of a backbone LLM, a set of task-specific PET modules, a PET module selector, and a cache system. We denote the PET module for the $k$-th task as $\mathrm{M}_k$. The working process at the $k$-th step can be summarized in two procedures: (1) PET module pre-selection (Section~\ref{sec:module-select}): We train the PET module selector to classify input examples into the $k$ seen schema sets $\{\mathcal{S}_1,\mathcal{S}_2,...,\mathcal{S}_k\}$. Then, $t$ PET modules corresponding to schema sets with top-$t$ selection scores are reserved as active ones. (2) Prediction with active PET modules (Section~\ref{sec:moe-predict}): Active PET modules produce logits on their own schema set respectively, which are then concatenated to make the final prediction. We analyze the training complexity in Section~\ref{sec:algorithm-complexity}. As an auxiliary module to address duplicate logit computations, the cache system is introduced in Section~\ref{sec:cache-system}. Dynamic ConPET also adopts the same dynamic replay strategy.

\subsubsection{PET Module Pre-Selection} \label{sec:module-select}

To avoid the uncontrollable linear growth of training costs as in former dynamic-architecture-based methods, we employ PET module pre-selection to select a fixed quantity of the most important PET modules. Specifically, at the $k$-th step, we train a PET module selector (also a PET module) to distinguish a fixed number $t$ of task schema sets that each example most probably belongs to among $\{\mathcal{S}_1,\mathcal{S}_2,...,\mathcal{S}_k\}$. Then the PET modules specific for the selected $t$ schema sets are reserved as active ones to participate in the subsequent inference.

Formally, the PET module selector converts each example $\mathbf{x}$ into a selection score vector $\mathbf{s}_{sel}$ of size $k$ following Formula~\ref{eq:encoding}, whose $j$-th element stands for the confidence that $\mathbf{x}$ belongs to $\mathcal{S}_j$. The top-$t$ elements of $\mathbf{s}_{sel}$ (with indexes $\{i_1,i_2,...,i_t\}$) determine the selection of $t$ active PET modules $\{\mathrm{M}_{i_1},\mathrm{M}_{i_2},...,\mathrm{M}_{i_t}\}$. Besides, we adopt the teacher-forcing policy that the correct PET module corresponding to the example is always selected for training. Through pre-selection, the forward propagation cost is kept unaffected by the number of PET modules. Experiments in Section~\ref{sec:effect-sel} also demonstrate its positive effect on performance.

\subsubsection{Prediction with Active PET Modules} \label{sec:moe-predict}

To obtain the final prediction, we integrate the learned information from active modules $\{\mathrm{M}_{i_1},\mathrm{M}_{i_2},...,\mathrm{M}_{i_t}\}$. Formally,
each active module obtains a logit vector on its corresponding schema set following Formula~\ref{eq:encoding}, while the logit vectors of inactive modules (i.e., those unselected modules) are assigned as vectors of an identical low enough value $\alpha$. We then concatenate the logits of all PET modules and figure out the prediction. This procedure can be expressed as the following formulas,

\begin{equation}
\begin{aligned}
    \label{eq:moe}
    &\mathbf{s}_j = f(\mathrm{M}_j, \textbf{x}),\ j\in\{i_1,i_2,...,i_t\}, \\
    &\mathbf{s}_j = [\alpha,\alpha,...,\alpha],\ j\notin\{i_1,i_2,...,i_t\}, \\
    &\mathbf{s}_j \in \mathbb{R}^{|\mathcal{S}_j|}, j=1,2,...,k, \\
    &pred = \arg\max\mathbf[\mathbf{s}_1,\mathbf{s}_2,...,\mathbf{s}_k], \\
\end{aligned}
\end{equation}

\noindent
where $[\cdot]$ denotes concatenation, $\mathbf{s}_j$ refers to the logit vector produced by the $j$-th PET module, and $pred$ means the predicted label.

\subsubsection{Detailed Algorithm and Complexity Analysis} \label{sec:algorithm-complexity}

Finally, we conduct an analysis of the detailed algorithm and its training complexity. The process of training Dynamic ConPET for the $k$-th task is presented in Algorithm~\ref{al:train-k}. Both the PET module selector and task-specific modules adopt a standard cross-entropy loss for multi-classification as the training objective. The PET module selector is iteratively updated. At the $k$-th step, the selector is initialized by the $(k-1)$-th step selector, except that its linear head should be expanded in dimension to handle more schemas, which is the function of $\mathrm{Dimension\_Expand}$.

\begin{algorithm}[ht]
    \caption{Dynamic ConPET for the $k$-th task}
    \label{al:train-k}
    \begin{algorithmic}[1]
        \REQUIRE The training set $\mathcal{T}_k$ of the $k$-th task
        \REQUIRE All the historical training data $\tilde{\mathcal{T}}_{k-1}$
        \REQUIRE All the seen task schema sets $\tilde{\mathcal{S}}_k$
        \REQUIRE The last PET module selector $\mathrm{M}^{(k-1)}_{sel}$
        \REQUIRE The maximum training batch number $iter_1$ and $iter_2$

        \STATE $\mathrm{M}^{(k)}_{sel} \leftarrow \mathrm{Dimension\_Expand}(\mathrm{M}^{(k-1)}_{sel})$
        \FOR{$i\leftarrow 1$ to $iter_1$}
            \STATE \textbf{Sample} $\mathcal{B}_{new}$ randomly from $\mathcal{T}_k$
            \STATE \textbf{Sample} $\mathcal{B}_{old}$ hierarchically from $\tilde{\mathcal{T}}_{k-1}$
            \STATE $\mathcal{B}_{tot} \leftarrow \mathcal{B}_{new} \cup \mathcal{B}_{old}$
            \STATE \textbf{Update} $\mathrm{M}^{(k)}_{sel}$ with $k$-category classification loss on $\mathcal{B}_{tot}$
        \ENDFOR

        \STATE \textbf{Initialize} the $k$-th task-specific module $\mathrm{M}_k$ 
        \FOR{$i\leftarrow 1$ to $iter_2$}
            \STATE \textbf{Sample} $\mathcal{B}_{new}$ randomly from $\mathcal{T}_k$
            \STATE \textbf{Sample} $\mathcal{B}_{old}$ hierarchically from $\tilde{\mathcal{T}}_{k-1}$
            \STATE $\mathcal{B}_{tot} \leftarrow \mathcal{B}_{new} \cup \mathcal{B}_{old}$
            \FOR{$\mathbf{x}$ in $\mathcal{B}_{tot}$}
                \STATE \textbf{Select} the most possible $t$ active PET modules for $\mathbf{x}$
                \STATE \textbf{Predict} for $\mathbf{x}$ with active PET modules
            \ENDFOR
            \STATE \textbf{Update} $\mathrm{M}_k$ with $|\tilde{\mathcal{S}}_k|$-category classification loss on $\mathcal{B}_{tot}$
        \ENDFOR
    \end{algorithmic}
\end{algorithm}

The above training process has a complexity independent of the task number. Specifically, we limit the training batch number of the PET module selector and the $k$-th task-specific module to $iter_1$ and $iter_2$ respectively, and the batch size is fixed as $b$. Meanwhile, $t$ active modules are reserved for each example. Therefore, the training complexity for the $k$-th task is $\mathcal{O}(b \cdot(iter_1 + iter_2 \cdot (t+1)))$, which does not increase with the task number $k$.

\subsubsection{Cache System} \label{sec:cache-system}

To address duplicate logit computations, we introduce an auxiliary cache system to store the logits generated by already-tuned PET modules. Specifically, each cache entry comprises the example ID, the PET module ID (the index of a task-specific module or the selector), and the logit vector. When a tuned PET module encounters an input example, ConPET checks the database using the example ID and PET module ID. If a match exists, the time-consuming calculation in Equation~\ref{eq:encoding} can be avoided. It should be noted that the cache system is available for a specific module only after the module completes its tuning process and remains fixed thereafter. For instance, the PET Module selector can only access the cache after its own training process in the pre-selection stage is finished.

\section{Experiments}

\begin{table*}[ht]
  \centering
  \setlength{\tabcolsep}{3.5mm}{
  \begin{threeparttable}
  \caption{The statistical information of 6 benchmarks involved in this work.}
  \begin{tabular}{l|c|c|c|c|c|c}
    \toprule
     & FewNERD & OntoNotes & BBN & FewRel & TACRED & ACE 2005 \\
    \midrule
    $\#$train & 340,383 & 211,523 & 89,365 & 44,800 & 13,012 & 5,648 \\
    $\#$valid & 48,759 & 26,440 & 11,150 & 5,600 & 5,436 & 700 \\
    $\#$test  & 96,902 & 26,441 & 11,213 & 5,600 & 3,325 & 722 \\
    $\#$total & 486,044 & 264,404 & 111,728 & 56,000 & 21,773 & 7,070 \\
    $\#$type & 66 & 86 & 46 & 80 & 41 & 18 \\
    $\#$cluster & 10 & 10 & 10 & 10 & 10 & 5 \\
    \bottomrule
  \end{tabular}
  \begin{tablenotes}
      \item ``$\#$train'', ``$\#$valid'', ``$\#$test'', and ``$\#$total'' refer to the number of examples in the training set, the validation set, the test set, and the whole dataset respectively. ``$\#$type'' represents the number of knowledge types contained in a dataset. ``$\#$cluster'' denotes the total task number $L$ of the corresponding benchmark.
  \end{tablenotes}
  \label{tab:data-stat}
  \end{threeparttable}
 }
\end{table*}

\begin{table*}[ht]
  \centering
  \setlength{\tabcolsep}{3.5mm}{
  \begin{threeparttable}
  \caption{The hyper-parameters used by ConPET on 6 datasets.}
  \begin{tabular}{l|c|c|c|c|c|c}
    \toprule
     & FewNERD & OntoNotes & BBN & FewRel & TACRED & ACE 2005 \\
    \midrule
    learning rate St-ConPET & $2e-5$ & $1e-4$ & $2e-4$ & $2e-4$ & $5e-4$ & $5e-4$ \\
    learning rate Dy-ConPET & $1e-4$ & $1e-4$ & $2e-4$ & $2e-4$ & $5e-4$ & $5e-4$ \\
    weight decay & 0.01 & 0.01 & 0.01 & 0.01 & 0.01 & 0.01 \\
    batch number limit & 25,000 & 12,500 & 5,000 & 4,000 & 2,000 & 2,000 \\
    training batch size  & 8 & 8 & 8 & 8 & 8 & 8 \\
    maximum epoch number & 10 & 10 & 10 & 10 & 20 & 20 \\
    maximum input length & 256 & 256 & 256 & 128 & 128 & 128 \\
    example number & 100 & 100 & 50 & 50 & 20 & 20 \\
    \bottomrule
  \end{tabular}
  \begin{tablenotes}
      \item ``St-ConPET'' and ``Dy-ConPET'' refer to the settings with Static ConPET and Dynamic ConPET respectively.
  \end{tablenotes}
  \label{tab:hyper-param}
  \end{threeparttable}
 }
\end{table*}

\subsection{Datasets}

For experiments, we focus on continual knowledge extraction as a representative continual adaptation scenario of LLMs, including entity typing and knowledge extraction. Specifically, we introduce the following 3 datasets as the benchmark for continual entity typing:

(1) \textbf{FewNERD}. FewNERD~\cite{ding2021few} is a large manually-labeled dataset with hierarchical 66 fine-grained entity types. Specifically, we use FewNERD (SUP) to construct the benchmark, including 486,044 examples from Wikipedia.

(2) \textbf{OntoNotes}. OntoNotes 5.0~\cite{weischedel2013ontonotes} is also a manually labeled dataset with 86 entity types, 264,404 sentences, and multiple sources of data.

(3) \textbf{BBN}. Compared with the above two datasets, BBN~\cite{weischedel2005bbn} is relatively smaller, with 111,728 examples and 46 entity types. It is mainly employed to test the generalizability of ConPET on small-scale tasks.

In addition, we test our method on 3 continual relation extraction tasks:

(1) \textbf{FewRel}. FewRel~\cite{han2018fewrel} is a large-scale dataset for relation extraction with 80 relation types and 56,000 examples from Wikipedia.

(2) \textbf{TACRED}. TACRED~\cite{zhang2017tacred} is a sentence-level relation extraction dataset obtained through crowdsourcing. For simplicity, we remove those examples with the label ``n/a'', and finally it consists of 21,773 sentences and 41 relation types.

(3) \textbf{ACE 2005}. We adopt the English part of ACE 2005 Multilingual Training Corpus~\cite{walker2006ace}, which includes 7,070 sentences and 18 relation types. Considering its small scale, it is also aimed at testing the small-scale-task generalizability of our method.

For each dataset, we construct the corresponding task sequence for continual learning by randomly splitting its entity or relation types into clusters (5 clusters for ACE 2005 and 10 clusters for the others). Then the schema set of each task is specified to one of these clusters.

The train-validation-test split methods of FewNERD and TACRED are the same as the original works~\cite{ding2021few,zhang2017tacred}. For the remaining 4 datasets, we randomly split them into the training, validation, and test set by a ratio of $8:1:1$. More key statistics of these 6 datasets are provided in Table~\ref{tab:data-stat}.

\subsection{Dataset Source and License} \label{sec:dataset-source}

All the textual materials included in the 6 datasets are in English. The examples of FewNERD and FewRel are collected from Wikipedia. The sentences in BBN are selected from the Penn Treebank Corpus of Wall Street Journal. The data sources of the remaining three datasets are mixed. TACRED is collected from the newswire and web collection. ACE 2005 contains textual materials from broadcast conversations, broadcast news, newsgroups, telephone conversations, and weblogs. OntoNotes 5.0 is the most complicated, with a corpus from a combination of telephone conversations, newswire, newsgroups, broadcast news, broadcast conversation, weblogs, and religious texts.

As for the licenses and data usage policy, FewNERD and FewRel are released under the CC BY-SA 4.0 license, while OntoNotes, BBN, TACRED, and ACE 2005 are used under the Linguistic Data Consortium (LDC) data license as a member. All the datasets are used in a way consistent with their intended use, and we limit public access to them under the requirements of licenses. Through manual sampling, we do not find any offensive content or identifiers in these datasets.

\begin{table*}[ht]
  \centering
  \setlength{\tabcolsep}{3mm}{
  \begin{threeparttable}
  \caption{The major experiment results on six datasets. ``W'' and ``A'' stand for the whole accuracy (\%) and average accuracy (\%) respectively. EMR*, EA-EMR*, and CRL* refer to the Static ConPET adapted version of EMR, EA-EMR, and CRL respectively. ``Dy-ConPET'' means Dynamic ConPET.}
  \begin{tabular}{l|cc|cc|cc|cc|cc|cc}
    \toprule
     & \multicolumn{2}{c|}{FewNERD} & \multicolumn{2}{c|}{OntoNotes} & \multicolumn{2}{c|}{BBN} & \multicolumn{2}{c|}{FewRel} & \multicolumn{2}{c|}{TACRED} & \multicolumn{2}{c}{ACE 2005}  \\
     & W & A & W & A & W & A & W & A & W & A & W & A \\
    \midrule
    \multicolumn{13}{c}{Baseline: Memory-based Methods $+$ PET Adaptation alone} \\
    \midrule
    EMR & 57.77 & 66.03 & 64.46 & 68.26 & 53.30 & 65.56 & 84.75 & 84.75 & 65.43 & 62.99 & 41.27 & 49.66 \\
    EA-EMR & 48.61 & 60.15 & 67.63 & 71.28 & 46.72 & 64.43 & 83.12 & 83.12 & 59.20 & 55.70 & 41.14 & 53.81 \\
    CRL & 72.73 & 75.48 & 61.39 & 62.65 & 55.98 & 64.02 & 84.38 & 84.37 & 75.91 & 72.61 & 54.29 & 61.96 \\
    \midrule
    \multicolumn{13}{c}{Memory-based Methods $+$ Static ConPET} \\
    \midrule
    EMR* & 74.41 & 77.13 & \textbf{85.12} & \textbf{86.11} & 80.05 & 83.13 & 89.30 & 89.30 & 80.44 & 75.75 & 79.85 & 80.53 \\
    EA-EMR* & 73.32 & 75.97 & 84.61 & 85.89 & \textbf{80.33} & \textbf{84.50} & 88.61 & 88.61 & 84.56 & 80.89 & \textbf{83.31} & \textbf{82.78} \\
    CRL* & 75.72 & 77.31 & 67.94 & 68.22 & 71.03 & 72.00 & \textbf{89.48} & \textbf{89.48} & \textbf{85.29} & \textbf{81.03} & 82.55 & 82.53 \\
    \midrule
    \multicolumn{13}{c}{Dynamic ConPET} \\
    \midrule
    Dy-ConPET & \textbf{76.15} & \textbf{78.22} & 84.47 & 85.83 & 76.15 & 81.16 & 88.62 & 88.62 & 84.47 & 80.27 & 80.27 & 80.76 \\
    \midrule
    \multicolumn{13}{c}{Upper Bound} \\
    \midrule
    Limitless & 79.49 & 77.80 & 88.19 & 87.73 & 89.61 & 85.97 & 90.34 & 90.34 & 87.40 & 82.58 & 84.07 & 83.22 \\
    \bottomrule
  \end{tabular}
  \begin{tablenotes}
    \small
    \item Bold entities represent the best results. Each result in this table except the time-consuming ``Limitless'' setting is averaged from a two-time replication.
  \end{tablenotes}
  \label{tab:overall-result}
  \end{threeparttable}
 }
\end{table*}

\begin{table*}[ht]
  \centering
  \setlength{\tabcolsep}{3.5mm}{
  \begin{threeparttable}
  \caption{The standard deviations of the major experimental results from two-time replication on 6 datasets.}
  \begin{tabular}{l|cc|cc|cc|cc|cc|cc}
    \toprule
     & \multicolumn{2}{c|}{FewNERD} & \multicolumn{2}{c|}{OntoNotes} & \multicolumn{2}{c|}{BBN} & \multicolumn{2}{c|}{FewRel} & \multicolumn{2}{c|}{TACRED} & \multicolumn{2}{c}{ACE 2005}  \\
     & W & A & W & A & W & A & W & A & W & A & W & A \\
    \midrule
    EMR & 2.87 & 1.66 & 0.67 & 0.50 & 2.41 & 1.28 & 0.70 & 0.70 & 0.49 & 0.51 & 0.55 & 0.17 \\
    EA-EMR & 3.32 & 3.33 & 1.90 & 1.73 & 2.12 & 0.34 & 2.17 & 2.17 & 0.07 & 0.41 & 1.11 & 1.22 \\
    CRL & 0.14 & 0.41 & 1.95 & 1.84 & 1.20 & 0.69 & 0.17 & 0.17 & 3.13 & 3.21 & 2.54 & 3.25 \\
    \midrule
    EMR* & 0.68 & 0.44 & 0.29 & 0.19 & 0.12 & 0.13 & 0.16 & 0.16 & 2.30 & 2.27 & 2.98 & 2.52 \\
    EA-EMR* & 1.61 & 0.12 & 0.73 & 0.81 & 0.53 & 0.10 & 0.52 & 0.52 & 0.91 & 0.52 & 1.18 & 2.10 \\
    CRL* & 0.01 & 0.06 & 0.53 & 0.51 & 0.34 & 0.05 & 0.07 & 0.07 & 0.33 & 0.58 & 0.97 & 1.27 \\
    \midrule
    Dy-ConPET & 0.12 & 0.01 & 0.02 & 0.02 & 1.66 & 0.34 & 0.57 & 0.57 & 0.34 & 0.66 & 1.17 & 1.02 \\
    \bottomrule
  \end{tabular}
  \begin{tablenotes}
      \item ``W'' and ``A'' stands for the whole accuracy (\%) and average accuracy (\%) respectively.
  \end{tablenotes}
  \label{tab:overall-std}
  \end{threeparttable}
 }
\end{table*}

\begin{figure*}[ht]
    \centering
    \includegraphics[width=0.8\linewidth]{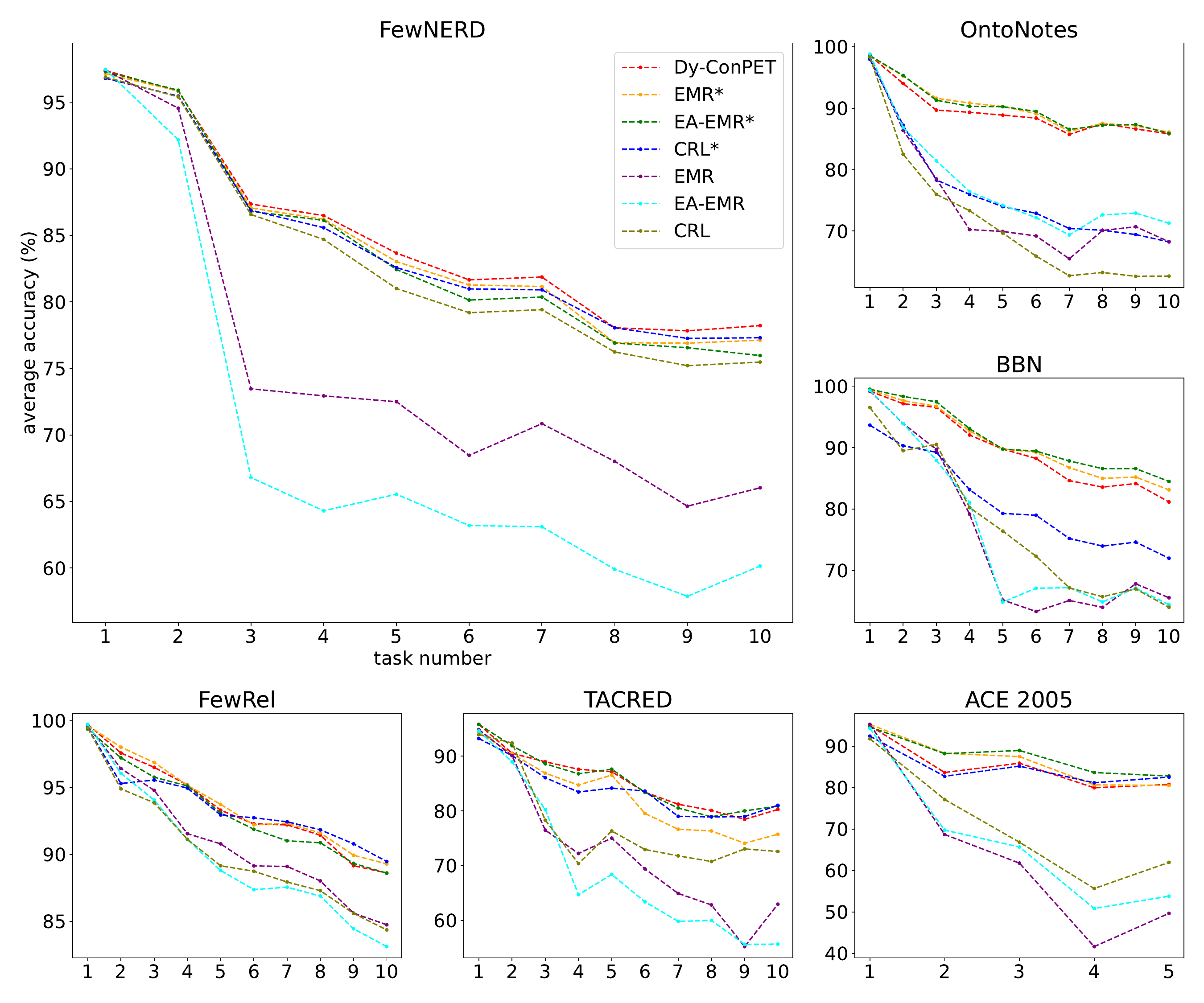}
    \caption{The average accuracies (\%) of different settings at each step throughout the learning process.}
    \label{fig:average-acc}
\end{figure*}

\subsection{Experimental Settings} \label{sec:exp-setting}

Two evaluation metrics are adopted: \textbf{whole accuracy} and \textbf{average accuracy}. The former is a standard classification accuracy on all the evaluation data $\tilde{\mathcal{Q}}_L$, where $L$ is the task number. The latter averages the independent accuracies of each seen task, which can better assess the ability to address catastrophic forgetting~\cite{han2020continual}.

We adopt LLaMA-7B~\cite{touvron2023llama} as the backbone LLM, which has 32 layers and a hidden size of 4,096. LoRA~\cite{hu2021lora} is used as a representative PET method in our experiments. Including LoRA matrices and a linear head, a PET module contains around 2M parameters, which only accounts for $0.03\%$ of the LLM (about 6.74B parameters). For efficiency, the rank of LoRA matrices and the number $t$ of active PET modules are set to 4 and 1 respectively under all settings. The ratio between old and new examples in each batch is fixed as $1:1$.

For Static ConPET, we mainly adapt it to the following three memory-based methods:

(1) \textbf{EMR}. EMR~\cite{parisi2019continual} is a basic memory-based continual learning method, which saves a fixed number of examples for each task and then replays them with new data combined when training a new model.

(2) \textbf{EA-EMR}. EA-EMR~\cite{wang2019sentence} is an extension of EMR. In addition to memory replay, it conducts embedding alignment at the end of each step to alleviate the distortion of the embedding space of old knowledge types.

(3) \textbf{CRL}. CRL~\cite{zhao2022consistent}, also an EMR extension, further applies contrastive
learning and knowledge distillation when replaying memorized examples to retain the old relation knowledge.

As the vanilla full-parameter fine-tuning occupies abundant computation resources with more than 3,000 times tunable parameters, we introduce the above three methods with PET adaptation alone as baselines, which adopt limited memory and select a fixed number of memorized examples for each knowledge type through K-Means clustering.

In addition, we introduce an upper-bound setting \textbf{Limitless} for reference. Specifically, it shares the same working process as Dynamic ConPET, except that its limit on the training batch number is removed. Therefore, at the $k$-th step, it is trained with full-volume historical data $\tilde{\mathcal{T}}_k$ and suffers from a linearly increasing complexity of $O(|\tilde{\mathcal{T}}_k|\cdot(t+2))$.

\subsection{Hyper-Parameters and Implementation Details}

The experimental hyper-parameters are tuned through grid search for each dataset respectively. Besides the most important parameters introduced in Section~\ref{sec:exp-setting}, other parameters for ConPET are listed in Table~\ref{tab:hyper-param}. Specifically, for Dynamic ConPET, the learning rate on FewNERD is adjusted to $5e-4$ for the first two continual learning steps, and the learning rate on OntoNotes is modified to $5e-4$ for the first three steps in all settings. The batch number limit refers to the restriction to the total training and validation batch number at each step if the maximum epoch number is reached, as mentioned in Section~\ref{sec:dynamic-replay}. The ratio between the training and validation batch number of ConPET is set to $4:1$ throughout our experiments. While the table provides the maximum epoch number, the best epoch is chosen according to the average accuracies on the validation data. Therefore, the actual training steps of the best-epoch model may be lower than the provided batch number limit. The ``example number'' refers to the memorized example number for each knowledge type at each step of the PET-only baseline version of three memory-based methods EMR, EA-EMR, and CRL, whose replay frequency is dynamically adjusted to obtain a total training step not lower than ConPET with dynamic sampling strategy.

The implementation of ConPET is based on PyTorch and each ConPET instance involved in our experiments is trained on a single A100 GPU for 1 to 48 hours, depending on the corresponding dataset scale. Tools and packages that we used in the experiments include: PyTorch, transformers, numpy, scikit-learn, tqdm, and loralib.

\subsection{Overall Results} \label{sec:overall-results}

The main experimental results are shown in Table~\ref{tab:overall-result}. The average accuracies of different settings at each step are shown in Figure~\ref{fig:average-acc}. From the table and figure, we can come to the following conclusions.

(1) Overall tendency: All the settings decline in average accuracies with the increase of task number, which reveals the inevitable impacts and challenges brought by catastrophic forgetting.

(2) Effectiveness: Both Static ConPET settings and Dynamic ConPET surpass the baselines by a large margin, which reveals their effectiveness. As all three memory-based methods adapted with Static ConPET significantly exceed corresponding PET-only versions, we can also demonstrate the importance of dynamic sampling strategy. Of course, gaps still exist between ConPET and the reference upper bound with linearly growing complexity, indicating that a large room remains for exploration in the continual task-specific fine-tuning of LLMs.

(3) Application Scenarios: While both Dynamic ConPET and Static ConPET yield satisfactory outcomes, the Static version excels on five benchmarks. A possible reason is the small scale of these datasets, which fits within the capacity of a single PET module, thus minimizing the negative impact of low scalability. Notably, Static ConPET demonstrates significant superiority in BBN and ACE 2005, which have the least knowledge schemas and examples among entity typing and relation extraction datasets respectively. In contrast, Dynamic ConPET outperforms Static ConPET on FewNERD with a large schema set and the most examples. Therefore, we can conclude that Static ConPET is more suitable for scenarios with relatively small-scale emerging data and knowledge schemas. Conversely, we need Dynamic ConPET to handle larger schema sets and more extensive data, which demand higher scalability of the continual fine-tuning architecture.


All the scores displayed in Table~\ref{tab:overall-result} except the ``Limitless'' setting are the average value of the results from a two-time replication with different random seeds. The standard deviations of these results are shown in Table~\ref{tab:overall-std}. Due to the extremely high complexity and training costs of ``Limitless'', we only take its result from a single run.

\subsection{Effect of Parameter-Efficient Tuning}  \label{sec:effect-pet}

\begin{table}[ht]
  \centering
  \setlength{\tabcolsep}{3.5mm}{
  \begin{threeparttable}
  \caption{The ablation results concerning the PET adaptation in ConPET based on BERT-Large and FewNERD benchmark.}
  \begin{tabular}{l|cc|cc}
    \toprule
     & \multicolumn{2}{c|}{PET} & \multicolumn{2}{c}{w/o PET} \\
     & W & A & W & A \\
    \midrule
    EMR* & 74.03 & 76.52 & 76.19 & 77.50 \\
    EA-EMR* & 74.59 & 77.96 & 77.64 & 75.90 \\
    CRL* & 73.80 & 75.87 & 77.29 & 77.70 \\
    Dy-ConPET & 74.25 & 76.60 & 76.97 & 76.87 \\
    \midrule
    $\#$param & \multicolumn{2}{c|}{\textbf{0.4M}} &\multicolumn{2}{c}{\textbf{335M}} \\
    \bottomrule
  \end{tabular}
  \begin{tablenotes}
      \item ``PET'' means the original ConPET setting and ``w/o PET'' refers to the parallel setting without PET. ``$\#$param'' is the number of tunable parameters.
  \end{tablenotes}
  \label{tab:ablation-pet}
  \end{threeparttable}
 }
\end{table}

Despite the fact that PET can significantly reduce the costs of parameter updates and GPU memory, it may lead to a decrease in overall performance~\cite{ding2022delta}. Due to the huge costs of tuning the full-parameter LLaMA-7B, we only involve the ablation study on the smaller 335M BERT-Large~\cite{devlin2018bert} as an alternative to demonstrating the rationality of PET adaptation. Concretely, we introduce the parallel settings on FewNERD for EMR*, EA-EMR*, CRL*, and Dynamic ConPET (Dy-ConPET) without PET adaptation. The results are shown in Table~\ref{tab:ablation-pet}. As can be observed, although ConPET has only 0.12\% tunable parameters and thus saves considerable time and computation resources, the drop in accuracy is not significant, which demonstrates the reasonableness of employing PET. The satisfactory results on BERT-Large also illustrate the generalizability of ConPET to relatively smaller pre-trained language models.

\subsection{Effect of PET Module Pre-Selection}  \label{sec:effect-sel}

\begin{table}[ht]
  \centering
  \setlength{\tabcolsep}{3.5mm}{
  \begin{threeparttable}
  \caption{The ablation results concerning the PET-module pre-selection in Dynamic ConPET on FewNERD and OntoNotes.}
  \begin{tabular}{l|cc|cc}
    \toprule
     & \multicolumn{2}{c|}{FewNERD} & \multicolumn{2}{c}{OntoNotes} \\
     & W & A & W & A \\
    \midrule
    Dy-ConPET & 76.15 & 78.22 & 84.47 & 85.83 \\
    w/o Sel & 70.75 & 76.22 & 81.85 & 83.37 \\
    \bottomrule
  \end{tabular}
  \begin{tablenotes}
      \item ``w/o Sel'' refers to the parallel setting without this technique.
  \end{tablenotes}
  \label{tab:ablation-sel}
  \end{threeparttable}
 }
\end{table}

In Section~\ref{sec:module-select}, we discuss the effect of PET module pre-selection on efficiency, which is to select a fixed number of active PET modules in Dynamic ConPET and ensure a constant forward propagation cost. Here we further analyze its effect from the aspect of performance. We conduct experiments on a parallel setting without this technique, which is corresponding to the situation with $t=k$ at the $k$-th step. The results are shown in Table~\ref{tab:ablation-sel}.

As can be observed, the performance significantly drops without PET module pre-selection, although the setting ``w/o Sel'' makes all PET modules active and has a linearly increasing complexity of $\mathcal{O}(b \cdot(iter_1 + iter_2 \cdot (k+1)))$. This may be attributed to the mutual interference between the logit vectors produced by $k$ PET modules. Even if the dynamic architecture can retain the classification capability on each independent task schema set, it is nontrivial to distinguish between the schemas of distinct tasks. Therefore, a PET module selector explicitly taking this responsibility can largely boost the overall performance as well as reduce the complexity.

\subsection{Effect of Different Task Splits} ~\label{sec:effect-split}

\begin{table}[ht]
  \centering
  \setlength{\tabcolsep}{3.5mm}{
  \begin{threeparttable}
  \caption{The performance of Dynamic ConPET on FewNERD given distinct task splits.}
  \begin{tabular}{l|ccc}
    \toprule
     & W & A & $\mathrm{Acc}_{sel}$ \\
    \midrule
    Correlated & 76.15 & 78.22 & 80.16 \\
    Independent & 77.62 & 76.18 & 88.91 \\
    \bottomrule
  \end{tabular}
  \begin{tablenotes}
      \item ``Correlated'' refers to the random split of length 10, while ``Independent'' is based on coarse types with length 8. $\mathrm{Acc}_{sel}$ denotes the top-1 accuracy of the PET module pre-selection.
  \end{tablenotes}
  \label{tab:ablation-split}
  \end{threeparttable}
 }
\end{table}

In our experiments, task splits are generated by randomly clustering knowledge schemas. However, these schemas may exhibit correlation, which leads to non-independent tasks. Such correlations have the potential to impact the performance of continual learning, especially Dynamic ConPET, whose architecture heavily relies on the task split. Therefore, this section will focus on the effect of different task splits on Dynamic ConPET.

Based on FewNERD, which has a hierarchical entity type schema with 8 coarse types, we reconstruct a new mutually independent task sequence of length 8 by assigning each coarse type to a task, while the original random split is inter-correlated. The results are shown in Table~\ref{tab:ablation-split}. While the overall accuracy does not change substantially, the independent split increases the pre-selection accuracy by a large margin, which implies a decrease in the accuracies of task-specific PET modules. This highlights a trade-off between the PET module pre-selection and downstream inner-task classification, as the upstream selector prefers an independent task split but the downstream PET modules favor a correlated one. To achieve optimal results, we should align the capacity of each PET module (including the selector) to its task difficulty. For instance, further splitting a task-specific PET module may be beneficial if the task exceeds its capacity.

Moreover, it may be reasonable to explore a wiser strategy for maintaining the task splits. A possible improvement is to adopt the knowledge-aware hierarchical organization of PET modules. Concretely, instead of limiting the layer number of PET modules to 2 (i.e., one layer of the PET selector and one layer of task-specific modules), we can introduce a PET module tree with multiple layers, where non-leaf PET modules are responsible for conducting pre-selection on its child nodes, and leaf PET modules make the final prediction. Meanwhile, considering the hierarchical nature of some knowledge schemas (e.g., the entity types in FewNERD, OntoNotes, and BBN), we can assign the responsibility of PET modules according to their positions in the knowledge hierarchy structure rather than the chronological order. Therefore, hierarchical knowledge can be incorporated explicitly and the capacity of each PET module can easily controlled given a well-designed knowledge schema. We leave this improvement for future work.

\section{Conclusion and Future Work}  \label{sec:conclusion}

In this paper, we mainly discuss the efficient and effective adaptation of LLMs to continual downstream task sequences. To achieve this goal, we propose the paradigm of ConPET, including two versions with training complexity independent of the task number. Static ConPET can adapt former memory-based methods to LLMs through the cost-saving PET and a dynamic sampling strategy more robust to over-fitting and forgetting. In contrast, Dynamic ConPET is more scalable to scenarios with large-scale data and task schemas owing to its dynamic MoE-style architecture. The experiments demonstrate the effectiveness and rationality of key techniques used in ConPET, with a considerable reduction in tuning costs. In the future, we will extend ConPET to more diverse continual learning scenarios (e.g., continual learning of heterogeneous tools~\cite{qin2023tool}) and further improve our paradigm by exploring wiser task split strategies.

{\appendix[More Supporting Facts for the Rationality of Dynamic Sampling Strategy] \label{sec:time_memory}


Despite the common assumption of limited memory in the field of continual learning~\cite{wang2019sentence}, we still consider it more reasonable to control the training costs rather than the memory during the process of fine-tuning LLMs, which is the fundamental philosophy of our dynamic sampling strategy. The rationality of this approach mainly lies in the overwhelming price of time and computation resources when compared to storage.

Take GPT-3 175B~\cite{brown2020language} as an example, whose training corpus contains about 300B tokens. As a single English token contains about 4 characters on average\footnote{\url{https://help.openai.com/en/articles/4936856-what-are-tokens-and-how-to-count-them}}, the overall storage for its training corpus is around $1\sim2$ TB. Such a scale of memory is quite acceptable for the majority of modern servers for AI research. Although the training materials for more recent LLMs are believed to take up more space, the price of storage has already been made moderate as no more than a few dozen dollars per TB in general thanks to the advance in storage hardware technology\footnote{\url{https://diskprices.com/}}. Besides, since the datasets of most downstream tasks should be filtered and annotated to ensure high quality and provide supervision for training, their scales can hardly reach the TB level and typically need much less storage than the LLM training corpus.

On the other hand, the computation resources required for GPT-3 175B (pre-trained on V100 GPUs) are at a more tremendous scale, which is around 3.64E+03 petaflop/s-days. Even if PET can save many computation resources, a single V100 GPU still costs thousands of dollars in a month\footnote{\url{https://cloud.google.com/compute/gpus-pricing}}, let alone the fact that LLMs with more than tens of billions of parameters may need more than one devices for fine-tuning and inference. In summary, the storage issue is just of little significance compared to the financial pressure posed by time and computation resources.
}

\bibliography{main}

\begin{thebibliography}{10}
\providecommand{\url}[1]{#1}
\csname url@samestyle\endcsname
\providecommand{\newblock}{\relax}
\providecommand{\bibinfo}[2]{#2}
\providecommand{\BIBentrySTDinterwordspacing}{\spaceskip=0pt\relax}
\providecommand{\BIBentryALTinterwordstretchfactor}{4}
\providecommand{\BIBentryALTinterwordspacing}{\spaceskip=\fontdimen2\font plus
\BIBentryALTinterwordstretchfactor\fontdimen3\font minus \fontdimen4\font\relax}
\providecommand{\BIBforeignlanguage}[2]{{%
\expandafter\ifx\csname l@#1\endcsname\relax
\typeout{** WARNING: IEEEtran.bst: No hyphenation pattern has been}%
\typeout{** loaded for the language `#1'. Using the pattern for}%
\typeout{** the default language instead.}%
\else
\language=\csname l@#1\endcsname
\fi
#2}}
\providecommand{\BIBdecl}{\relax}
\BIBdecl

\bibitem{brown2020language}
\BIBentryALTinterwordspacing
T.~Brown, B.~Mann, N.~Ryder, M.~Subbiah, J.~D. Kaplan, P.~Dhariwal, A.~Neelakantan, P.~Shyam, G.~Sastry, A.~Askell \emph{et~al.}, ``Language models are few-shot learners,'' \emph{Advances in neural information processing systems}, vol.~33, pp. 1877--1901, 2020. [Online]. Available: \url{https://proceedings.neurips.cc/paper_files/paper/2020/file/1457c0d6bfcb4967418bfb8ac142f64a-Paper.pdf}
\BIBentrySTDinterwordspacing

\bibitem{wei2021finetuned}
\BIBentryALTinterwordspacing
J.~Wei, M.~Bosma, V.~Y. Zhao, K.~Guu, A.~W. Yu, B.~Lester, N.~Du, A.~M. Dai, and Q.~V. Le, ``Finetuned language models are zero-shot learners,'' \emph{arXiv preprint arXiv:2109.01652}, 2021. [Online]. Available: \url{https://arxiv.org/pdf/2109.01652}
\BIBentrySTDinterwordspacing

\bibitem{ouyang2022training}
\BIBentryALTinterwordspacing
L.~Ouyang, J.~Wu, X.~Jiang, D.~Almeida, C.~Wainwright, P.~Mishkin, C.~Zhang, S.~Agarwal, K.~Slama, A.~Ray \emph{et~al.}, ``Training language models to follow instructions with human feedback,'' \emph{Advances in Neural Information Processing Systems}, vol.~35, pp. 27\,730--27\,744, 2022. [Online]. Available: \url{https://proceedings.neurips.cc/paper_files/paper/2022/file/b1efde53be364a73914f58805a001731-Paper-Conference.pdf}
\BIBentrySTDinterwordspacing

\bibitem{ding2022delta}
\BIBentryALTinterwordspacing
N.~Ding, Y.~Qin, G.~Yang, F.~Wei, Z.~Yang, Y.~Su, S.~Hu, Y.~Chen, C.-M. Chan, W.~Chen \emph{et~al.}, ``Delta tuning: A comprehensive study of parameter efficient methods for pre-trained language models,'' \emph{arXiv preprint arXiv:2203.06904}, 2022. [Online]. Available: \url{https://arxiv.org/pdf/2203.06904.pdf}
\BIBentrySTDinterwordspacing

\bibitem{jin2022lifelong}
\BIBentryALTinterwordspacing
X.~Jin, D.~Zhang, H.~Zhu, W.~Xiao, S.-W. Li, X.~Wei, A.~Arnold, and X.~Ren, ``Lifelong pretraining: Continually adapting language models to emerging corpora,'' in \emph{Proceedings of the 2022 Conference of the North American Chapter of the Association for Computational Linguistics: Human Language Technologies}, 2022, pp. 4764--4780. [Online]. Available: \url{https://aclanthology.org/2022.naacl-main.351.pdf}
\BIBentrySTDinterwordspacing

\bibitem{daruna2021continual}
\BIBentryALTinterwordspacing
A.~Daruna, M.~Gupta, M.~Sridharan, and S.~Chernova, ``Continual learning of knowledge graph embeddings,'' \emph{IEEE Robotics and Automation Letters}, vol.~6, no.~2, pp. 1128--1135, 2021. [Online]. Available: \url{https://ieeexplore.ieee.org/abstract/document/9343669}
\BIBentrySTDinterwordspacing

\bibitem{monaikul2021continual}
\BIBentryALTinterwordspacing
N.~Monaikul, G.~Castellucci, S.~Filice, and O.~Rokhlenko, ``Continual learning for named entity recognition,'' in \emph{Proceedings of the AAAI Conference on Artificial Intelligence}, 2021, pp. 13\,570--13\,577. [Online]. Available: \url{https://ojs.aaai.org/index.php/AAAI/article/view/17600/17407}
\BIBentrySTDinterwordspacing

\bibitem{qin2023tool}
\BIBentryALTinterwordspacing
Y.~Qin, S.~Hu, Y.~Lin, W.~Chen, N.~Ding, G.~Cui, Z.~Zeng, Y.~Huang, C.~Xiao, C.~Han \emph{et~al.}, ``Tool learning with foundation models,'' \emph{arXiv preprint arXiv:2304.08354}, 2023. [Online]. Available: \url{https://arxiv.org/pdf/2304.08354}
\BIBentrySTDinterwordspacing

\bibitem{zhao2022consistent}
\BIBentryALTinterwordspacing
K.~Zhao, H.~Xu, J.~Yang, and K.~Gao, ``Consistent representation learning for continual relation extraction,'' in \emph{Findings of the Association for Computational Linguistics: ACL 2022}, 2022, pp. 3402--3411. [Online]. Available: \url{https://aclanthology.org/2022.findings-acl.268.pdf}
\BIBentrySTDinterwordspacing

\bibitem{parisi2019continual}
\BIBentryALTinterwordspacing
G.~I. Parisi, R.~Kemker, J.~L. Part, C.~Kanan, and S.~Wermter, ``Continual lifelong learning with neural networks: A review,'' \emph{Neural Networks}, vol. 113, pp. 54--71, 2019. [Online]. Available: \url{https://www.sciencedirect.com/science/article/pii/S0893608019300231}
\BIBentrySTDinterwordspacing

\bibitem{rusu2016progressive}
\BIBentryALTinterwordspacing
A.~A. Rusu, N.~C. Rabinowitz, G.~Desjardins, H.~Soyer, J.~Kirkpatrick, K.~Kavukcuoglu, R.~Pascanu, and R.~Hadsell, ``Progressive neural networks,'' \emph{arXiv preprint arXiv:1606.04671}, 2016. [Online]. Available: \url{https://arxiv.org/pdf/1606.04671.pdf}
\BIBentrySTDinterwordspacing

\bibitem{gu2021transformer}
\BIBentryALTinterwordspacing
X.~Gu, L.~Liu, H.~Yu, J.~Li, C.~Chen, and J.~Han, ``On the transformer growth for progressive {BERT} training,'' in \emph{Proceedings of the 2021 Conference of the North American Chapter of the Association for Computational Linguistics: Human Language Technologies}, 2021, pp. 5174--5180. [Online]. Available: \url{https://aclanthology.org/2021.naacl-main.406.pdf}
\BIBentrySTDinterwordspacing

\bibitem{rolnick2019experience}
\BIBentryALTinterwordspacing
D.~Rolnick, A.~Ahuja, J.~Schwarz, T.~Lillicrap, and G.~Wayne, ``Experience replay for continual learning,'' \emph{Advances in Neural Information Processing Systems}, vol.~32, 2019. [Online]. Available: \url{https://proceedings.neurips.cc/paper/2019/file/fa7cdfad1a5aaf8370ebeda47a1ff1c3-Paper.pdf}
\BIBentrySTDinterwordspacing

\bibitem{wang2019sentence}
\BIBentryALTinterwordspacing
H.~Wang, W.~Xiong, M.~Yu, X.~Guo, S.~Chang, and W.~Y. Wang, ``Sentence embedding alignment for lifelong relation extraction,'' in \emph{Proceedings of the 2019 Conference of the North American Chapter of the Association for Computational Linguistics: Human Language Technologies, Volume 1 (Long and Short Papers)}, 2019, pp. 796--806. [Online]. Available: \url{https://aclanthology.org/N19-1086.pdf}
\BIBentrySTDinterwordspacing

\bibitem{han2020continual}
\BIBentryALTinterwordspacing
X.~Han, Y.~Dai, T.~Gao, Y.~Lin, Z.~Liu, P.~Li, M.~Sun, and J.~Zhou, ``Continual relation learning via episodic memory activation and reconsolidation,'' in \emph{Proceedings of the 58th Annual Meeting of the Association for Computational Linguistics}, 2020, pp. 6429--6440. [Online]. Available: \url{https://aclanthology.org/2020.acl-main.573.pdf}
\BIBentrySTDinterwordspacing

\bibitem{devlin2018bert}
\BIBentryALTinterwordspacing
J.~Devlin, M.-W. Chang, K.~Lee, and K.~Toutanova, ``{BERT}: Pre-training of deep bidirectional transformers for language understanding,'' \emph{arXiv preprint arXiv:1810.04805}, 2018. [Online]. Available: \url{https://arxiv.org/pdf/1810.04805.pdf}
\BIBentrySTDinterwordspacing

\bibitem{liu2019roberta}
\BIBentryALTinterwordspacing
Y.~Liu, M.~Ott, N.~Goyal, J.~Du, M.~Joshi, D.~Chen, O.~Levy, M.~Lewis, L.~Zettlemoyer, and V.~Stoyanov, ``{RoBERTa}: A robustly optimized bert pretraining approach,'' \emph{arXiv preprint arXiv:1907.11692}, 2019. [Online]. Available: \url{https://arxiv.org/pdf/1907.11692.pdf}
\BIBentrySTDinterwordspacing

\bibitem{houlsby2019parameter}
\BIBentryALTinterwordspacing
N.~Houlsby, A.~Giurgiu, S.~Jastrzebski, B.~Morrone, Q.~De~Laroussilhe, A.~Gesmundo, M.~Attariyan, and S.~Gelly, ``Parameter-efficient transfer learning for {NLP},'' in \emph{International Conference on Machine Learning}.\hskip 1em plus 0.5em minus 0.4em\relax PMLR, 2019, pp. 2790--2799. [Online]. Available: \url{http://proceedings.mlr.press/v97/houlsby19a/houlsby19a.pdf}
\BIBentrySTDinterwordspacing

\bibitem{li2021prefix}
\BIBentryALTinterwordspacing
X.~L. Li and P.~Liang, ``{Prefix-Tuning}: Optimizing continuous prompts for generation,'' in \emph{Proceedings of the 59th Annual Meeting of the Association for Computational Linguistics and the 11th International Joint Conference on Natural Language Processing (Volume 1: Long Papers)}, 2021, pp. 4582--4597. [Online]. Available: \url{https://aclanthology.org/2021.acl-long.353.pdf}
\BIBentrySTDinterwordspacing

\bibitem{gao2021making}
\BIBentryALTinterwordspacing
T.~Gao, A.~Fisch, and D.~Chen, ``Making pre-trained language models better few-shot learners,'' in \emph{Proceedings of the 59th Annual Meeting of the Association for Computational Linguistics and the 11th International Joint Conference on Natural Language Processing (Volume 1: Long Papers)}, 2021, pp. 3816--3830. [Online]. Available: \url{https://aclanthology.org/2021.acl-long.295.pdf}
\BIBentrySTDinterwordspacing

\bibitem{lee2019would}
\BIBentryALTinterwordspacing
J.~Lee, R.~Tang, and J.~Lin, ``What would elsa do? freezing layers during transformer fine-tuning,'' \emph{arXiv preprint arXiv:1911.03090}, 2019. [Online]. Available: \url{https://arxiv.org/pdf/1911.03090.pdf}
\BIBentrySTDinterwordspacing

\bibitem{zhao2020masking}
\BIBentryALTinterwordspacing
M.~Zhao, T.~Lin, F.~Mi, M.~Jaggi, and H.~Sch{\"u}tze, ``Masking as an efficient alternative to finetuning for pretrained language models,'' in \emph{Proceedings of the 2020 Conference on Empirical Methods in Natural Language Processing}, 2020, pp. 2226--2241. [Online]. Available: \url{https://aclanthology.org/2020.emnlp-main.174.pdf}
\BIBentrySTDinterwordspacing

\bibitem{zaken2022bitfit}
\BIBentryALTinterwordspacing
E.~B. Zaken, Y.~Goldberg, and S.~Ravfogel, ``Bitfit: Simple parameter-efficient fine-tuning for transformer-based masked language-models,'' in \emph{Proceedings of the 60th Annual Meeting of the Association for Computational Linguistics (Volume 2: Short Papers)}, 2022, pp. 1--9. [Online]. Available: \url{https://aclanthology.org/2022.acl-short.1.pdf}
\BIBentrySTDinterwordspacing

\bibitem{hu2021lora}
\BIBentryALTinterwordspacing
E.~J. Hu, Y.~Shen, P.~Wallis, Z.~Allen-Zhu, Y.~Li, S.~Wang, L.~Wang, and W.~Chen, ``{LoRA}: Low-rank adaptation of large language models,'' \emph{arXiv preprint arXiv:2106.09685}, 2021. [Online]. Available: \url{https://arxiv.org/pdf/2106.09685.pdf}
\BIBentrySTDinterwordspacing

\bibitem{qin2021exploring}
\BIBentryALTinterwordspacing
Y.~Qin, X.~Wang, Y.~Su, Y.~Lin, N.~Ding, Z.~Liu, J.~Li, L.~Hou, P.~Li, M.~Sun \emph{et~al.}, ``Exploring low-dimensional intrinsic task subspace via prompt tuning,'' \emph{arXiv preprint arXiv:2110.07867}, 2021. [Online]. Available: \url{https://arxiv.org/pdf/2110.07867.pdf}
\BIBentrySTDinterwordspacing

\bibitem{sun2023comparative}
\BIBentryALTinterwordspacing
X.~Sun, Y.~Ji, B.~Ma, and X.~Li, ``A comparative study between full-parameter and lora-based fine-tuning on chinese instruction data for instruction following large language model,'' \emph{arXiv preprint arXiv:2304.08109}, 2023. [Online]. Available: \url{https://arxiv.org/pdf/2304.08109.pdf}
\BIBentrySTDinterwordspacing

\bibitem{touvron2023llama}
\BIBentryALTinterwordspacing
H.~Touvron, T.~Lavril, G.~Izacard, X.~Martinet, M.-A. Lachaux, T.~Lacroix, B.~Rozi{\`e}re, N.~Goyal, E.~Hambro, F.~Azhar \emph{et~al.}, ``{LLaMA}: Open and efficient foundation language models,'' \emph{arXiv preprint arXiv:2302.13971}, 2023. [Online]. Available: \url{https://arxiv.org/pdf/2302.13971.pdf}
\BIBentrySTDinterwordspacing

\bibitem{kirkpatrick2017overcoming}
\BIBentryALTinterwordspacing
J.~Kirkpatrick, R.~Pascanu, N.~Rabinowitz, J.~Veness, G.~Desjardins, A.~A. Rusu, K.~Milan, J.~Quan, T.~Ramalho, A.~Grabska-Barwinska \emph{et~al.}, ``Overcoming catastrophic forgetting in neural networks,'' \emph{Proceedings of the national academy of sciences}, vol. 114, no.~13, pp. 3521--3526, 2017. [Online]. Available: \url{https://www.pnas.org/doi/epdf/10.1073/pnas.1611835114}
\BIBentrySTDinterwordspacing

\bibitem{lee2017overcoming}
\BIBentryALTinterwordspacing
S.-W. Lee, J.-H. Kim, J.~Jun, J.-W. Ha, and B.-T. Zhang, ``Overcoming catastrophic forgetting by incremental moment matching,'' in \emph{Proceedings of the 31st International Conference on Neural Information Processing Systems}, 2017, pp. 4655--4665. [Online]. Available: \url{https://proceedings.neurips.cc/paper/2017/file/f708f064faaf32a43e4d3c784e6af9ea-Paper.pdf}
\BIBentrySTDinterwordspacing

\bibitem{chaudhry2018riemannian}
\BIBentryALTinterwordspacing
A.~Chaudhry, P.~K. Dokania, T.~Ajanthan, and P.~H. Torr, ``Riemannian walk for incremental learning: Understanding forgetting and intransigence,'' in \emph{Proceedings of the European Conference on Computer Vision (ECCV)}, 2018, pp. 532--547. [Online]. Available: \url{https://openaccess.thecvf.com/content_ECCV_2018/papers/Arslan_Chaudhry__Riemannian_Walk_ECCV_2018_paper.pdf}
\BIBentrySTDinterwordspacing

\bibitem{li2017learning}
\BIBentryALTinterwordspacing
Z.~Li and D.~Hoiem, ``Learning without forgetting,'' \emph{IEEE transactions on pattern analysis and machine intelligence}, vol.~40, no.~12, pp. 2935--2947, 2017. [Online]. Available: \url{https://ieeexplore.ieee.org/abstract/document/8107520}
\BIBentrySTDinterwordspacing

\bibitem{zhang2020class}
\BIBentryALTinterwordspacing
J.~Zhang, J.~Zhang, S.~Ghosh, D.~Li, S.~Tasci, L.~Heck, H.~Zhang, and C.-C.~J. Kuo, ``Class-incremental learning via deep model consolidation,'' in \emph{Proceedings of the IEEE/CVF Winter Conference on Applications of Computer Vision}, 2020, pp. 1131--1140. [Online]. Available: \url{https://openaccess.thecvf.com/content_WACV_2020/papers/Zhang_Class-incremental_Learning_via_Deep_Model_Consolidation_WACV_2020_paper.pdf}
\BIBentrySTDinterwordspacing

\bibitem{chen2015net2net}
\BIBentryALTinterwordspacing
T.~Chen, I.~Goodfellow, and J.~Shlens, ``{Net2Net}: Accelerating learning via knowledge transfer,'' \emph{arXiv preprint arXiv:1511.05641}, 2015. [Online]. Available: \url{https://arxiv.org/pdf/1511.05641.pdf}
\BIBentrySTDinterwordspacing

\bibitem{isele2018selective}
\BIBentryALTinterwordspacing
D.~Isele and A.~Cosgun, ``Selective experience replay for lifelong learning,'' in \emph{Proceedings of the AAAI Conference on Artificial Intelligence}, 2018. [Online]. Available: \url{https://ojs.aaai.org/index.php/AAAI/article/view/11595/11454}
\BIBentrySTDinterwordspacing

\bibitem{qin2021lfpt5}
\BIBentryALTinterwordspacing
C.~Qin and S.~Joty, ``{LFPT5}: A unified framework for lifelong few-shot language learning based on prompt tuning of {T5},'' in \emph{International Conference on Learning Representations}, 2021. [Online]. Available: \url{https://openreview.net/pdf?id=HCRVf71PMF}
\BIBentrySTDinterwordspacing

\bibitem{razdaibiedina2022progressive}
\BIBentryALTinterwordspacing
A.~Razdaibiedina, Y.~Mao, R.~Hou, M.~Khabsa, M.~Lewis, and A.~Almahairi, ``Progressive prompts: Continual learning for language models,'' in \emph{The Eleventh International Conference on Learning Representations}, 2022. [Online]. Available: \url{https://openreview.net/pdf?id=UJTgQBc91_}
\BIBentrySTDinterwordspacing

\bibitem{madotto2021continual}
\BIBentryALTinterwordspacing
A.~Madotto, Z.~Lin, Z.~Zhou, S.~Moon, P.~A. Crook, B.~Liu, Z.~Yu, E.~Cho, P.~Fung, and Z.~Wang, ``Continual learning in task-oriented dialogue systems,'' in \emph{Proceedings of the 2021 Conference on Empirical Methods in Natural Language Processing}, 2021, pp. 7452--7467. [Online]. Available: \url{https://aclanthology.org/2021.emnlp-main.590.pdf}
\BIBentrySTDinterwordspacing

\bibitem{gao2023unified}
\BIBentryALTinterwordspacing
Q.~Gao, C.~Zhao, Y.~Sun, T.~Xi, G.~Zhang, B.~Ghanem, and J.~Zhang, ``A unified continual learning framework with general parameter-efficient tuning,'' \emph{arXiv preprint arXiv:2303.10070}, 2023. [Online]. Available: \url{https://arxiv.org/pdf/2303.10070.pdf}
\BIBentrySTDinterwordspacing

\bibitem{qin2022elle}
\BIBentryALTinterwordspacing
Y.~Qin, J.~Zhang, Y.~Lin, Z.~Liu, P.~Li, M.~Sun, and J.~Zhou, ``Elle: Efficient lifelong pre-training for emerging data,'' in \emph{Findings of the Association for Computational Linguistics: ACL 2022}, 2022, pp. 2789--2810. [Online]. Available: \url{https://aclanthology.org/2022.findings-acl.220.pdf}
\BIBentrySTDinterwordspacing

\bibitem{ke2023continual}
\BIBentryALTinterwordspacing
Z.~Ke, Y.~Shao, H.~Lin, T.~Konishi, G.~Kim, and B.~Liu, ``Continual pre-training of language models,'' in \emph{The Eleventh International Conference on Learning Representations}, 2023. [Online]. Available: \url{https://openreview.net/pdf?id=m_GDIItaI3o}
\BIBentrySTDinterwordspacing

\bibitem{shazeer2017outrageously}
\BIBentryALTinterwordspacing
N.~Shazeer, A.~Mirhoseini, K.~Maziarz, A.~Davis, Q.~Le, G.~Hinton, and J.~Dean, ``Outrageously large neural networks: The sparsely-gated mixture-of-experts layer,'' in \emph{International Conference on Learning Representations}, 2017. [Online]. Available: \url{https://openreview.net/pdf?id=B1ckMDqlg}
\BIBentrySTDinterwordspacing

\bibitem{lepikhin2020gshard}
\BIBentryALTinterwordspacing
D.~Lepikhin, H.~Lee, Y.~Xu, D.~Chen, O.~Firat, Y.~Huang, M.~Krikun, N.~Shazeer, and Z.~Chen, ``{GShard}: Scaling giant models with conditional computation and automatic sharding,'' \emph{arXiv preprint arXiv:2006.16668}, 2020. [Online]. Available: \url{https://arxiv.org/pdf/2006.16668}
\BIBentrySTDinterwordspacing

\bibitem{lewis2021base}
\BIBentryALTinterwordspacing
M.~Lewis, S.~Bhosale, T.~Dettmers, N.~Goyal, and L.~Zettlemoyer, ``{BASE Layers}: Simplifying training of large, sparse models,'' in \emph{International Conference on Machine Learning}.\hskip 1em plus 0.5em minus 0.4em\relax PMLR, 2021, pp. 6265--6274. [Online]. Available: \url{http://proceedings.mlr.press/v139/lewis21a/lewis21a.pdf}
\BIBentrySTDinterwordspacing

\bibitem{roller2021hash}
\BIBentryALTinterwordspacing
S.~Roller, S.~Sukhbaatar, J.~Weston \emph{et~al.}, ``Hash layers for large sparse models,'' \emph{Advances in Neural Information Processing Systems}, vol.~34, pp. 17\,555--17\,566, 2021. [Online]. Available: \url{https://proceedings.neurips.cc/paper_files/paper/2021/file/92bf5e6240737e0326ea59846a83e076-Paper.pdf}
\BIBentrySTDinterwordspacing

\bibitem{fedus2022switch}
\BIBentryALTinterwordspacing
W.~Fedus, B.~Zoph, and N.~Shazeer, ``{Switch Transformers}: Scaling to trillion parameter models with simple and efficient sparsity,'' \emph{The Journal of Machine Learning Research}, vol.~23, no.~1, pp. 5232--5270, 2022. [Online]. Available: \url{https://jmlr.org/papers/volume23/21-0998/21-0998.pdf}
\BIBentrySTDinterwordspacing

\bibitem{zhang2022moefication}
\BIBentryALTinterwordspacing
Z.~Zhang, Y.~Lin, Z.~Liu, P.~Li, M.~Sun, and J.~Zhou, ``{MoEfication}: Transformer feed-forward layers are mixtures of experts,'' in \emph{Findings of the Association for Computational Linguistics: ACL 2022}, 2022, pp. 877--890. [Online]. Available: \url{https://aclanthology.org/2022.findings-acl.71.pdf}
\BIBentrySTDinterwordspacing

\bibitem{gururangan2022demix}
\BIBentryALTinterwordspacing
S.~Gururangan, M.~Lewis, A.~Holtzman, N.~A. Smith, and L.~Zettlemoyer, ``{DEMix} layers: Disentangling domains for modular language modeling,'' in \emph{Proceedings of the 2022 Conference of the North American Chapter of the Association for Computational Linguistics: Human Language Technologies}, 2022, pp. 5557--5576. [Online]. Available: \url{https://aclanthology.org/2022.naacl-main.407.pdf}
\BIBentrySTDinterwordspacing

\bibitem{shen2023mixtureofexperts}
\BIBentryALTinterwordspacing
S.~Shen, L.~Hou, Y.~Zhou, N.~Du, S.~Longpre, J.~Wei, H.~W. Chung, B.~Zoph, W.~Fedus, X.~Chen, T.~Vu, Y.~Wu, W.~Chen, A.~Webson, Y.~Li, V.~Zhao, H.~Yu, K.~Keutzer, T.~Darrell, and D.~Zhou, ``Mixture-of-experts meets instruction tuning: A winning combination for large language models,'' \emph{arXiv preprint arXiv:2305.14705}, 2023. [Online]. Available: \url{https://arxiv.org/pdf/2305.14705.pdf}
\BIBentrySTDinterwordspacing

\bibitem{zadouri2023pushing}
\BIBentryALTinterwordspacing
T.~Zadouri, A.~{\"U}st{\"u}n, A.~Ahmadian, B.~Ermi{\c{s}}, A.~Locatelli, and S.~Hooker, ``Pushing mixture of experts to the limit: Extremely parameter efficient moe for instruction tuning,'' \emph{arXiv preprint arXiv:2309.05444}, 2023. [Online]. Available: \url{https://arxiv.org/pdf/2309.05444.pdf}
\BIBentrySTDinterwordspacing

\bibitem{soares2019matching}
\BIBentryALTinterwordspacing
L.~B. Soares, N.~Fitzgerald, J.~Ling, and T.~Kwiatkowski, ``Matching the blanks: Distributional similarity for relation learning,'' in \emph{Proceedings of the 57th Annual Meeting of the Association for Computational Linguistics}, 2019, pp. 2895--2905. [Online]. Available: \url{https://aclanthology.org/P19-1279.pdf}
\BIBentrySTDinterwordspacing

\bibitem{ding2021prompt}
\BIBentryALTinterwordspacing
N.~Ding, Y.~Chen, X.~Han, G.~Xu, P.~Xie, H.-T. Zheng, Z.~Liu, J.~Li, and H.-G. Kim, ``Prompt-learning for fine-grained entity typing,'' \emph{arXiv preprint arXiv:2108.10604}, 2021. [Online]. Available: \url{https://arxiv.org/pdf/2108.10604.pdf}
\BIBentrySTDinterwordspacing

\bibitem{lester2021power}
\BIBentryALTinterwordspacing
B.~Lester, R.~Al-Rfou, and N.~Constant, ``The power of scale for parameter-efficient prompt tuning,'' in \emph{Proceedings of the 2021 Conference on Empirical Methods in Natural Language Processing}, 2021, pp. 3045--3059. [Online]. Available: \url{https://aclanthology.org/2021.emnlp-main.243.pdf}
\BIBentrySTDinterwordspacing

\bibitem{ding2021few}
\BIBentryALTinterwordspacing
N.~Ding, G.~Xu, Y.~Chen, X.~Wang, X.~Han, P.~Xie, H.~Zheng, and Z.~Liu, ``{Few-NERD}: A few-shot named entity recognition dataset,'' in \emph{Proceedings of the 59th Annual Meeting of the Association for Computational Linguistics and the 11th International Joint Conference on Natural Language Processing (Volume 1: Long Papers)}, 2021, pp. 3198--3213. [Online]. Available: \url{https://aclanthology.org/2021.acl-long.248.pdf}
\BIBentrySTDinterwordspacing

\bibitem{weischedel2013ontonotes}
\BIBentryALTinterwordspacing
R.~Weischedel, M.~Palmer, M.~Marcus, E.~Hovy, S.~Pradhan, L.~Ramshaw, N.~Xue, A.~Taylor, J.~Kaufman, M.~Franchini \emph{et~al.}, ``{OntoNotes} release 5.0 ldc2013t19,'' \emph{Linguistic Data Consortium, Philadelphia, PA}, vol.~23, 2013. [Online]. Available: \url{https://catalog.ldc.upenn.edu/LDC2013T19}
\BIBentrySTDinterwordspacing

\bibitem{weischedel2005bbn}
\BIBentryALTinterwordspacing
R.~Weischedel and A.~Brunstein, ``{BBN} pronoun coreference and entity type corpus,'' \emph{Linguistic Data Consortium, Philadelphia}, vol. 112, 2005. [Online]. Available: \url{https://catalog.ldc.upenn.edu/LDC2005T33}
\BIBentrySTDinterwordspacing

\bibitem{han2018fewrel}
\BIBentryALTinterwordspacing
X.~Han, H.~Zhu, P.~Yu, Z.~Wang, Y.~Yao, Z.~Liu, and M.~Sun, ``{FewRel}: A large-scale supervised few-shot relation classification dataset with state-of-the-art evaluation,'' in \emph{Proceedings of the 2018 Conference on Empirical Methods in Natural Language Processing}, 2018, pp. 4803--4809. [Online]. Available: \url{https://aclanthology.org/D18-1514.pdf}
\BIBentrySTDinterwordspacing

\bibitem{zhang2017tacred}
\BIBentryALTinterwordspacing
Y.~Zhang, V.~Zhong, D.~Chen, G.~Angeli, and C.~D. Manning, ``Position-aware attention and supervised data improve slot filling,'' in \emph{Proceedings of the 2017 Conference on Empirical Methods in Natural Language Processing}, 2017, pp. 35--45. [Online]. Available: \url{https://nlp.stanford.edu/pubs/zhang2017tacred.pdf}
\BIBentrySTDinterwordspacing

\bibitem{walker2006ace}
\BIBentryALTinterwordspacing
C.~Walker, S.~Strassel, J.~Medero, and K.~Maeda, ``{ACE} 2005 multilingual training corpus,'' \emph{Linguistic Data Consortium, Philadelphia}, vol.~57, p.~45, 2006. [Online]. Available: \url{https://catalog.ldc.upenn.edu/LDC2006T06}
\BIBentrySTDinterwordspacing

\end{thebibliography}
\bibliographystyle{IEEEtran}


 





\end{document}